\documentclass[letterpaper]{article} 

\ifdefined\aaaianonymous
    \usepackage[submission]{aaai2026}  
\else
    \usepackage{aaai2026}              
\fi

\usepackage{times}  
\usepackage{helvet}  
\usepackage{courier}  
\usepackage[hyphens]{url}  
\usepackage{graphicx} 
\usepackage{natbib}  
\usepackage{caption} 
\usepackage{algorithm}
\usepackage{algorithmic}
\usepackage{tabularx}
\usepackage{tcolorbox} 
\usepackage{subcaption}
\usepackage{booktabs}
\usepackage{enumitem}
\usepackage{pifont}
\usepackage{amsmath,amssymb,amsfonts}

\usepackage{newfloat}
\usepackage{listings}
\DeclareCaptionStyle{ruled}{labelfont=normalfont,labelsep=colon,strut=off} 
\floatstyle{ruled}
\newfloat{listing}{tb}{lst}{}
\floatname{listing}{Listing}

\ifdefined\aaaianonymous
    \title{ToC: Tree-of-Claims Search with Multi-Agent Language Models}
\else
    \title{ToC: Tree-of-Claims Search with Multi-Agent Language Models}
\fi

\title{ToC: Tree-of-Claims Search with Multi-Agent Language Models}
\author {
    Shuyang Yu\textsuperscript{\rm 1}\thanks{Work done during an internship at AIGROW.}, Jianan Liang\textsuperscript{\rm 2}, Hui Hu\textsuperscript{\rm 2}
}
\affiliations {
    \textsuperscript{\rm 1}Columbia University\\
    \textsuperscript{\rm 2}AIGROW\\
    sy3309@columbia.edu,liangjianan@aigrow.cn,huhui@aigrow.cn
}

\usepackage{bibentry}

\begin{document}

\maketitle

\begin{abstract}
Optimizing patent claims is a critical yet challenging task, demanding careful balance between maximizing novelty and preserving legal scope. Manual claim drafting is labor-intensive, costly, and inherently inconsistent, while conventional Large Language Models (LLMs) often lack the structured, iterative reasoning essential for precise claim refinement. To address these challenges, we introduce Tree of Claims (ToC), an innovative framework that redefines claim editing as a guided search problem. ToC synergistically integrates Monte Carlo Tree Search (MCTS) with a collaborative multi-agent system, comprising an LLM-based EditorAgent that proposes contextually grounded edits, and an ExaminerAgent that mimics patent examiner critiques through structured, chain-of-thought analyses of novelty and prior art disclosure. Driven by a carefully designed multi-objective reward function, ToC jointly optimizes novelty, scope retention, and semantic coherence. Experimental evaluation on a benchmark of 1145 claims demonstrates that ToC significantly outperforms standard LLMs in zero-shot and few-shot scenarios, achieving an average composite score improvement of 8\%, and up to 9\% in certain cases. Extensive experiments, including detailed ablation studies, validate ToC's efficacy in generating superior, legally robust claim revisions. Overall, ToC establishes a transparent, controllable, and interpretable methodology that effectively bridges advanced LLM reasoning capabilities with strategic MCTS planning for structured patent claim optimization. The source code is available at https://github.com/ysy2003/ToC.
\end{abstract}

\ifdefined\aaaianonymous
\else
\fi

\section{Introduction}

Drafting and revising patent claims is crucial, significantly influencing the legal scope, technical breadth, and commercial potential of intellectual property \cite{son2022ai, jiang2024can, kawano2024claimbrush}. When addressing examiner rejections or seeking to optimize claim language for stronger protection, practitioners must navigate complex demands: ensuring precise legal wording, maintaining technical coverage, and maximizing novelty over prior art \cite{paul2024revolutionary}. Conventionally, these tasks are executed by experienced patent professionals who rely heavily on domain expertise, iterative refinement, and sophisticated understanding of examiner criteria. However, manual processes are resource-intensive, costly, and inherently inconsistent \cite{wang2024patentformer}. Consequently, a critical question emerges: can we construct a systematic, controllable, and transparent approach that effectively integrates the generative capabilities of large language models (LLMs) with structured planning methods to automate and enhance patent claim drafting?

Building on notable progress in pre‑trained models, including foundation models~\cite{fang2025neuript} and LLMs~\cite{fang2025see}, across domains such as medicine, law, and education, recent advancements have increasingly enabled the automation of patent-related tasks. Nevertheless, existing LLM-driven solutions predominantly operate under single-shot or few-shot paradigms, lacking the iterative and structured reasoning crucial for high-quality claim engineering \cite{wang2024autopatent, ren2024patentgpt, chu2024paris, bai2024patentgpt}. These approaches often inadequately balance the intricate interplay among legal scope, technical specificity, and linguistic precision, thus restricting their effectiveness in practical scenarios.

Addressing these critical shortcomings, we introduce the Tree of Claims (ToC), an innovative framework that reconceptualizes patent claim editing as a structured search process. ToC integrates Monte Carlo Tree Search (MCTS) with a sophisticated multi-agent collaboration mechanism. Specifically, an LLM-based EditorAgent proposes contextually accurate, legally sound edits, while an ExaminerAgent simulates detailed patent examiner scrutiny, providing structured assessments of novelty and disclosure.

The ToC framework employs an uncertainty-aware MCTS architecture, wherein each node represents an ordered sequence of discrete editing operations. Optimization is guided by a carefully crafted multi-objective reward function, considering novelty enhancement, scope preservation, and semantic coherence. Additional techniques such as uncertainty gating and progressive widening ensure claim improvements remain both interpretable and verifiable.

The core contributions of this paper are:
\begin{itemize}
\item A novel formulation of patent claim optimization as a structured search problem within the innovative ToC framework, integrating multi-agent collaborative reasoning with MCTS.
\item A collaborative multi-agent architecture where an EditorAgent generates legally robust edit operations, and an ExaminerAgent provides systematic assessments, jointly guiding a targeted, multi-objective search process.
\item Extensive experiments demonstrating that ToC significantly outperforms state-of-the-art LLM-based baselines (including zero-shot and few-shot settings) in novelty, scope, and legal robustness across real-world patent revision tasks.
\end{itemize}

\section{Related Work}
Patent claim optimization lies at the confluence of structured text editing, legal-domain reasoning, and sequential decision-making. We review related work across three core areas: LLMs for patent drafting and revision, Multi-agent systems for structured reasoning and Search-based text optimization.

\paragraph{LLMs for Patent Drafting and Revision.}
LLMs like GPT-4 are increasingly used in patent tasks, from claim drafting to responding to office actions \cite{son2022ai,chu2024paris}. However, these black-box models offer limited control over output structure, scope, or legal consistency \cite{li2024control}. Their one-shot drafts often require substantial human revision, failing to bridge the gap between technical details and legal coverage. While domain-specific fine-tuning improves fluency, it doesn't solve the core controllability issue \cite{ren2024patentgpt,bai2024patentgpt}. Even top LLMs necessitate iterative human refinement for full legal robustness, and recent benchmarks highlight persistent gaps in claim quality \cite{jiang2024can}. Legal-theory work also warns that generative AI can inadvertently expand claim scope without proper safeguards \cite{wang2024prompts}.

Researchers are now pursuing controllable, feedback-driven approaches. ClaimBrush \cite{kawano2024claimbrush} incorporates examiner feedback but remains largely a one-pass revision. Other iterative methods, like AutoPatent \cite{wang2024autopatent} and EvoPat \cite{wang2024evopat}, use multi-agent frameworks, but their editing operations are often opaque and non-deterministic. This lack of transparency hinders real-world integration, where practitioners need to verify every change. To address this, the ToC framework provides systematic, stepwise claim edits with precise control and multi-criteria feedback. Each modification in ToC is explicit and justified, ensuring traceable revisions and verifiable legal scope at every step—crucial for high-stakes patent applications.

\paragraph{Multi-Agent Systems for Structured Reasoning.}
Multi-agent LLM frameworks are powerful for complex tasks, enabling specialized agents to collaborate and critique each other for more coherent results. Examples include MetaGPT \cite{hong2023metagpt}, AutoGen \cite{wu2024autogen}, OpenAgents \cite{xie2023openagents}, LongWriter \cite{bai2024longwriter}, and PARIS-style collaborative systems for patent responses \cite{chu2024paris}. These structured approaches enhance LLM reasoning through iterative self-correction.

However, most existing multi-agent systems are for open-ended tasks without strict domain constraints. Patent claim editing, in contrast, demands fine-grained control and strict adherence to legal criteria \cite{bui2025advancing}. ToC excels here by embedding these constraints directly into its agent roles: an EditorAgent proposes controlled, atomic, legally-grounded edits, while an ExaminerAgent provides structured, fine-grained feedback on each edit. This interactive critique loop ensures a transparent, justified revision history. Crucially, ToC's agents operate under strict domain-specific objectives, explicitly checking compliance with patent rules. Unlike prior systems that yield only a final text, ToC generates a complete, reasoned edit history, providing an interpretable and auditable chain of revisions vital for legal practitioners.

\paragraph{Search-Based Text Optimization.}
Search algorithms offer a powerful way to enhance generative LLMs, especially for tasks requiring detailed planning or adherence to multiple constraints. MCTS has been revitalized for text generation through speculative parallelization \cite{cheng2024speculative} and dynamic-memory guidance \cite{shi2025monte}. MCTS also underpins frameworks like Tree-of-Thoughts \cite{yao2023tree}, which explore multiple reasoning paths for complex problem-solving. These advances complement earlier evolutionary or RL-based strategies \cite{wu2025boosting}, yielding more controllable and reliable text generation.

Applying search to legal text, particularly patent claims, presents unique challenges due to strict requirements for coherence, legal compliance, and scope preservation.  ToC addresses this by integrating an uncertainty‑aware MCTS with specialized LLM agents \cite{hu2024uncertainty,cheng2024speculative}. It defines atomic edit operations as moves in the search tree, ensuring every revision path is plausible \cite{agrawal2021non}.  A multi‑objective reward function guides the MCTS, balancing novelty gain against scope and legal coherence \cite{yuan2024self}. A novel $\sigma$‑gating mechanism further filters candidate edits: the ExaminerAgent reports a variance‑based confidence score, and moves with $\sigma$ above a preset threshold are pruned, reducing erroneous commitments \cite{ling2024uncertainty}. These innovations make ToC’s search process controllable and interpretable, providing a traceable and auditable chain of revisions critical for high‑stakes patent claim editing.

\section{Methodology}
Recent advances in LLMs have significantly impacted legal text generation but often fall short in high-stakes editing tasks like patent claim revision. Single-shot or few-shot models generally fail to incorporate essential components such as legal constraints, strategic reasoning, and iterative refinement required for expert-level patent drafting.

Observing real-world workflows, we identify two primary roles employed by patent professionals: generating legally valid edits and evaluating these edits for novelty, clarity, and scope. To systematically manage their interaction and effectively control the extensive editing space, we model the claim optimization problem as a structured search task.

\subsection{Problem Formulation}
Given an initial claim $C_0$ and a set of prior art documents $P = \{P_1, \dots, P_m\}$, the objective is to generate a revised claim $C_T$ that maximizes novelty over $P$ while preserving the technical scope of $C_0$. The process is modeled as a sequential decision-making problem where each state $s_t$ represents a claim configuration at time step $t$, and each action $a_t$ represents an atomic editing operation. Additionally, the prior art documents may include images that provide contextual information relevant to the claim.

The action space $\mathcal{A}$ includes ten atomic operations. Each operation $a_t$ is characterized by a tuple $(o_t, e_t, r_t, c_t)$, where $o_t$ is the operation type, $e_t$ is the target element identifier, $r_t$ is the reasoning chain, and $c_t$ is the confidence score. See Table~\ref{tab:edit-ops} for a summary of atomic operations.

To account for editing dependencies and ensure semantic validity, we define the precedence relations among actions. For instance, $\textsc{AddNovelFeature}$ must precede $\textsc{ReplaceSynonym}$ to ensure semantic coherence.

The goal is to find an optimal sequence $\mathcal{A}^* = (a_0, \dots, a_{T-1})$ that maximizes the expected cumulative reward:
\begin{equation}
\mathcal{A}^* = \arg\max_{\mathcal{A}} \mathbb{E}\left[\sum_{t=0}^{T-1} R(s_t, a_t)\right]
\label{eq:optimal-sequence}
\end{equation}
where $R(s_t, a_t)$ is the reward function that evaluates the quality of the modification at each step.

\begin{table}[ht]
\centering
\begin{tabularx}{\linewidth}{lX}
\toprule
\textbf{Operation} & \textbf{Description} \\
\midrule
AddNovelFeature & Introduce a new technical feature \\
ReplaceSynonym & Replace with a synonym \\
ReframeViaFigure & Reframe using a figure \\
DropElement & Remove an element \\
MergeElements & Merge two or more elements \\
SplitElement & Split an element into parts \\
AddLimitation & Add a limiting condition \\
ModifyRelationship & Modify the relationship between elements \\
ChangeOrder & Change the order of elements \\
AddDependency & Add a dependency between elements \\
\bottomrule
\end{tabularx}
\caption{Summary of Atomic Edit Operations in ToC.}
\label{tab:edit-ops}
\end{table}

\subsection{ToC Search Architecture}

\begin{figure}[ht]
    \centering
    \includegraphics[width=1.0\columnwidth]{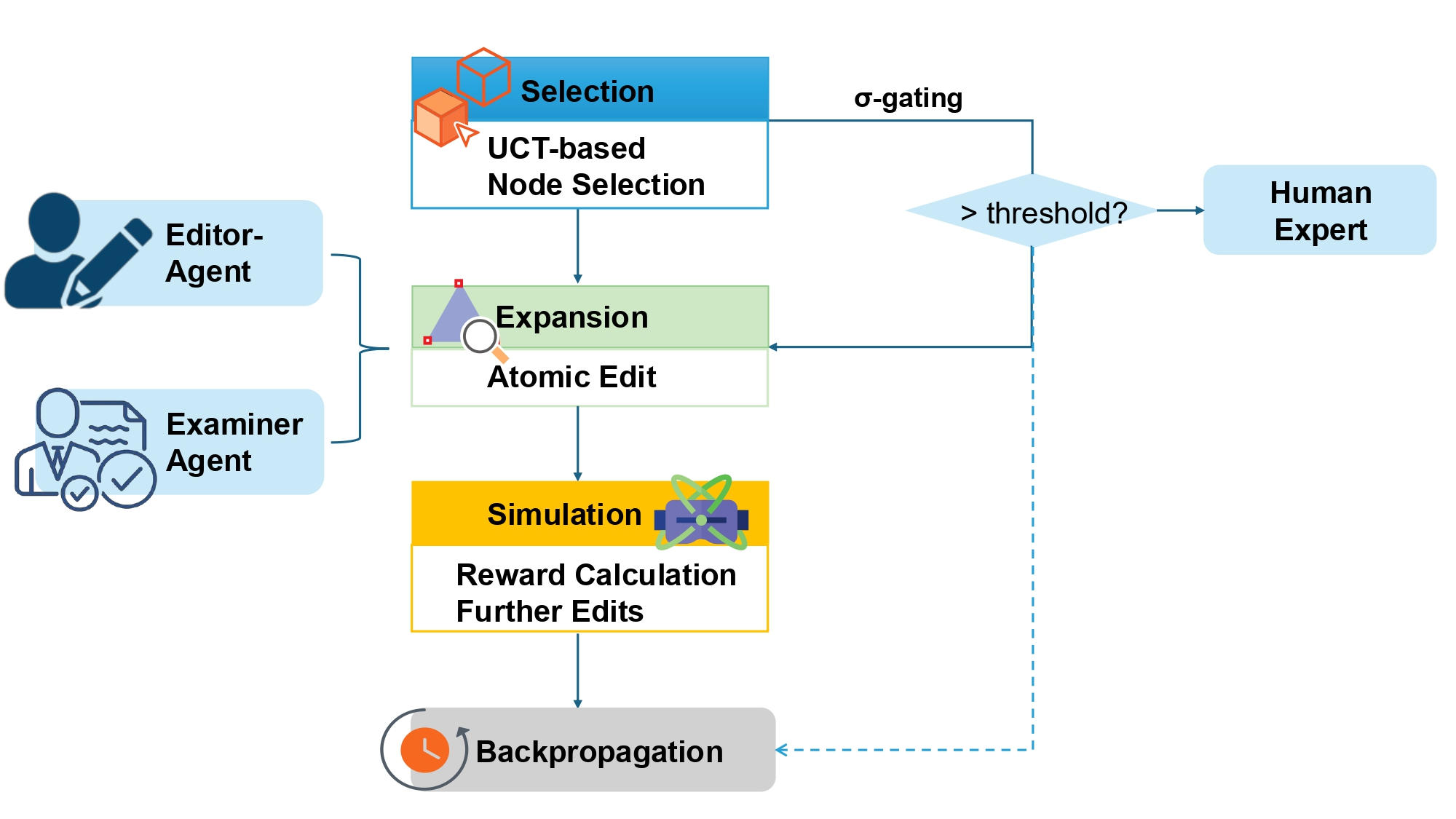}
    \caption{Detailed workflow of the ToC methodology.}
    \label{fig:toc_pipeline}
\end{figure}

In addressing the problem of patent claim optimization, we propose a new search architecture to effectively generate revised claims. Figure~\ref{fig:toc_pipeline} presents the comprehensive ToC workflow, integrating Monte Carlo Tree Search (MCTS), multi-agent collaboration, and uncertainty-aware controls. The search traverses four phases: Selection, Expansion, Simulation, and Backpropagation, leveraging uncertainty gating with a threshold ($\sigma_{epi}^{max}=0.2$) determined through rigorous tuning experiments optimizing performance and consistency.

The search process is organized as follows: Each node in the search tree represents a claim state $n = (C, O, R, \sigma)$, where $C$ is the claim text, $O$ is the edit operation sequence, $R$ is cumulative reward, and $\sigma = (\sigma^{\text{epi}}, \sigma^{\text{ale}})$ represents epistemic and aleatoric uncertainty components. The tree is explored through four enhanced phases:

\begin{enumerate}[label=\arabic*., noitemsep, topsep=0pt]

\item \textbf{Selection.}  
Starting at the root, the search descends by repeatedly choosing the child with the highest UCT score
\begin{equation}
\text{UCT}(n)=\frac{Q(n)}{N(n)}+c\sqrt{\frac{\ln N(p)}{N(n)}},
\label{eq:uct-formula}
\end{equation}
where $Q(n)$ and $N(n)$ are the cumulative reward and visit count of node~$n$, $N(p)$ is the visit count of its parent, and $c$ controls exploration.  
\textbf{$\sigma$‑Gating.}  For each visited node we estimate epistemic variance $\sigma_{\text{epi}}(n)$.  Paths with $\sigma_{\text{epi}}(n)>\sigma^{\text{epi}}_{\max}=0.2$ are pruned or flagged for human review, steering the search away from highly uncertain regions.

\item \textbf{Expansion.}  
Once an expandable node is reached, \textsc{ExaminerAgent} pinpoints candidate claim elements; \textsc{EditorAgent} then proposes concrete atomic edits.  To keep the branching factor manageable we apply \emph{progressive widening}
\begin{equation}
K(n)=\bigl\lceil\alpha\, N(n)^{\delta}\bigr\rceil,\qquad \alpha>0,\;0<\delta<1,
\label{eq:progressive-widening-K}
\end{equation}
so that only the first $K(n)$ high‑value edits become children, ensuring semantic validity while controlling tree growth.

\item \textbf{Simulation.}  
From the newly added child we roll out a complete claim using a rule‑guided policy with three selectable modes:  
\emph{entropy‑based} (favor edits with high uncertainty),  
\emph{confidence‑based} (favor edits with high agent confidence), and  
\emph{hybrid} (weighted mix $0.6/0.4$).

\item \textbf{Backpropagation.}  
The reward $R(C_t)$ of the simulated claim $C_t$ is propagated up the path:  
\begin{equation}
Q(n)\leftarrow Q(n)+R(C_t),\qquad
N(n)\leftarrow N(n)+1.
\label{eq:backprop}
\end{equation}

\end{enumerate}

The search terminates when any of the following conditions are met: maximum iterations $T_{\text{max}} = 800$ reached, reward improvement stalls ($|R_t - R_{t-1}| < \epsilon = 0.01$), maximum search time $T_{\text{search}} = 3600$ seconds exceeded, or consecutive failures exceed threshold $N_{\text{fail}} = 20$.

\begin{algorithm}
\caption{ToC-MCTS with Uncertainty-Aware Search}
\label{alg:toc-mcts}
\begin{algorithmic}[1]
\STATE Initialize root node $N_0 \leftarrow (C_0, \emptyset, 0, \sigma_0)$
\STATE Initialize uncertainty decomposer and progressive widening
\FOR{$t = 1$ to $T_{\text{max}}$}
    \STATE $N_{\text{sel}} \leftarrow \text{Select}(N_0)$
    \IF{$\sigma_{\text{epi}}(N_{\text{sel}}) > \sigma_{\text{max}}^{\text{epi}}$}
        \STATE Trigger human intervention or strategy switch
    \ENDIF
    \STATE $N_{\text{exp}} \leftarrow \text{Expand}(N_{\text{sel}})$
    \STATE $R \leftarrow \text{Simulate}(N_{\text{exp}})$
    \STATE \text{Backprop}($N_{\text{exp}}$, $R$)
    \IF{$\text{ShouldTerminate}()$}
        \STATE \textbf{break}
    \ENDIF
\ENDFOR
\RETURN best $C^*$ on highest reward path
\end{algorithmic}
\end{algorithm}

\subsection{Multi-Agent Collaboration}
The ToC framework employs a collaborative pair of specialized LLM-driven agents that work in concert to achieve optimal claim modification:

\subsubsection{Examiner Agent}

The \textsc{ExaminerAgent} performs disclosure analysis by examining each claim element against prior art documents, generating structured reasoning chains $\mathcal{R} = \{r_1, \dots, r_n\}$. Each reasoning chain $r_i$ contains:

\begin{itemize}
    \item \textbf{Status:} $\text{Disclosed}$, $\text{NotDisclosed}$, or $\text{PartiallyDisclosed}$
    \item \textbf{Evidence text:} Direct quotes from prior art supporting the judgment
    \item \textbf{Confidence score:} $c_i \in [0,1]$ indicating the examiner's confidence
    \item \textbf{Uncertainty measure:} $\sigma_i \in [0,1]$ quantifying decision uncertainty
    \item \textbf{Human intervention flag:} $h_i \in \{0,1\}$ indicating need for human review
\end{itemize}

The examiner agent uses a specialized prompt template that requires strict JSON output format to ensure consistent parsing and evaluation. When uncertainty exceeds the threshold $\sigma_{\text{max}}^{\text{epi}} = 0.2$, the system flags the case for human intervention.

\subsubsection{Editor Agent}

The \textsc{EditorAgent} generates atomic edit operations based on the examiner's feedback, applying operations that transform disclosed elements while preserving claim scope. The agent maintains operation history and evaluates modification quality through multiple metrics:

\begin{itemize}
    \item \textbf{Text similarity:} Measures semantic preservation between original and modified claims
    \item \textbf{Scope preservation:} Quantifies the degree to which the original claim scope is maintained
    \item \textbf{Legal readability:} Assesses whether the modified claim maintains patent-legal language standards
    \item \textbf{Technical coherence:} Evaluates the logical consistency of the modified claim
\end{itemize}

The editor agent employs a sophisticated operation selection strategy that considers both the disclosure status of elements and the potential impact of each operation on claim scope and novelty.

This collaborative loop ensures that each edit is both legally robust and technically meaningful. The concise prompt templates for both agents are summarized in Table~\ref{tab:prompt_templates_concise}.

\begin{table}[ht]
\centering
\small
\renewcommand{\arraystretch}{1.1}
\begin{tabularx}{\linewidth}{@{}p{2.6cm}X@{}}
\toprule
\textbf{Agent Role} & \textbf{Core Task and Key Instructions} \\
\midrule
\textbf{Examiner Agent} &
Analyzes a claim element against prior art.
\begin{itemize}[leftmargin=1em, noitemsep]
    \item Decompose into sub-points.
    \item Check each for disclosure (synonym/functional/structural equivalence).
    \item Output strict JSON:
    \texttt{claim\_points}, \texttt{matched\_prior\_art}, \texttt{judgement}, \texttt{reasoning}.
    \item No natural language outside JSON.
\end{itemize}
Inputs: \texttt{<claim\_element>}, \texttt{<prior\_art>} \\
\midrule
\textbf{Editor Agent} &
Revises claims based on examiner feedback to enhance novelty.
\begin{itemize}[leftmargin=1em, noitemsep]
    \item Focus on “disclosed” sub-points; preserve “not disclosed” parts.
    \item Avoid trivial paraphrasing; introduce technical novelty.
    \item Maintain grammar and logical coherence.
    \item Output only the revised claim text.
\end{itemize}
Inputs: \texttt{Original claim}, \texttt{Judgement chain} \\
\bottomrule
\end{tabularx}
\caption{Summarized prompt templates for Examiner Agent and Editor Agent.}
\label{tab:prompt_templates_concise}
\end{table}

\begin{table*}[t]
\centering
\small
\begin{tabular}{llcccccc}
\toprule
\textbf{Model} & \textbf{ToC} & $R_\mathrm{cov}$ & $R_\mathrm{scope}$ & $R_\mathrm{novelty}$ & $R_\mathrm{cons}$ & $R_\mathrm{uncert}$ & \textbf{Overall} \\
\midrule
\multicolumn{8}{l}{\textbf{Closed‑Source MLLMs}}\\
OpenAI O1                 & \ding{51} & 0.560±0.048 & 0.374±0.051 & 0.712±0.050 & 0.942±0.012 & 0.071±0.014 & 0.680±0.03 \\
\hspace{2.7em}+ zero‑shot & \ding{55} & 0.492±0.044 & 0.401±0.053 & 0.643±0.048 & 0.931±0.019 & 0.079±0.020 & 0.631±0.03 \\
\hspace{2.7em}+ few‑shot  & \ding{55} & 0.525±0.037 & 0.388±0.056 & 0.685±0.041 & 0.937±0.018 & 0.075±0.020 & 0.658±0.03 \\[2pt]
GPT‑4o                    & \ding{51} & 0.582±0.050 & 0.389±0.050 & 0.732±0.050 & 0.956±0.017 & 0.068±0.018 & \textbf{0.701±0.03} \\
\hspace{2.7em}+ zero‑shot & \ding{55} & 0.520±0.046 & 0.417±0.052 & 0.659±0.049 & 0.947±0.018 & 0.071±0.019 & 0.647±0.03 \\
\hspace{2.7em}+ few‑shot  & \ding{55} & 0.555±0.045 & 0.405±0.051 & 0.698±0.047 & 0.951±0.014 & 0.070±0.019 & 0.678±0.03 \\[2pt]
Claude‑3.5 Sonnet         & \ding{51} & 0.548±0.049 & 0.370±0.057 & 0.703±0.051 & 0.945±0.018 & 0.075±0.020 & 0.675±0.03 \\
\hspace{2.7em}+ zero‑shot & \ding{55} & 0.472±0.045 & 0.408±0.053 & 0.633±0.047 & 0.934±0.019 & 0.082±0.021 & 0.626±0.03 \\
\hspace{2.7em}+ few‑shot  & \ding{55} & 0.510±0.044 & 0.388±0.052 & 0.652±0.046 & 0.940±0.018 & 0.080±0.022 & 0.653±0.03 \\[2pt]
\midrule
\multicolumn{8}{l}{\textbf{Open‑Source MLLMs}}\\
Qwen2.5‑VL‑32B            & \ding{51} & 0.507±0.047 & 0.351±0.053 & 0.665±0.050 & 0.924±0.027 & 0.083±0.025 & 0.639±0.03 \\
\hspace{2.7em}+ zero‑shot & \ding{55} & 0.424±0.043 & 0.382±0.057 & 0.598±0.049 & 0.913±0.028 & 0.090±0.026 & 0.592±0.03 \\
\hspace{2.7em}+ few‑shot  & \ding{55} & 0.468±0.044 & 0.370±0.051 & 0.635±0.048 & 0.918±0.028 & 0.087±0.025 & 0.615±0.03 \\[2pt]
Qwen2.5‑VL‑72B            & \ding{51} & 0.534±0.048 & 0.361±0.056 & 0.682±0.051 & 0.930±0.026 & 0.080±0.024 & 0.658±0.03 \\
\hspace{2.7em}+ zero‑shot & \ding{55} & 0.452±0.044 & 0.392±0.054 & 0.624±0.048 & 0.921±0.020 & 0.088±0.025 & 0.611±0.03 \\
\hspace{2.7em}+ few‑shot  & \ding{55} & 0.495±0.045 & 0.378±0.051 & 0.655±0.047 & 0.926±0.027 & 0.084±0.024 & 0.636±0.03 \\
\bottomrule
\end{tabular}
\caption{Primary evaluation metrics for ToC and zero‑/few‑shot baselines on five MLLMs (mean $\pm$ SD over 3 seeds, $N=500$).}
\label{tab:llm_primary_metrics}
\end{table*}

\subsection{Reward Function Design}
The Monte Carlo tree must balance five often‑competing objectives—novelty coverage, scope retention, technical novelty, logical consistency, and risk control—within a single scalar score.  
We therefore combine them linearly
\begin{equation}
\label{eq:reward-function}
\begin{aligned}
R(C_t)=&\;w_1\,R_{\text{coverage}}(C_t)-w_2\,R_{\text{scope}}(C_t)\\
      &+w_3\,R_{\text{novelty}}(C_t)+w_4\,R_{\text{consistency}}(C_t)\\
      &-w_5\,R_{\text{uncertainty}}(C_t),
\end{aligned}
\end{equation}

where $(w_1,\ldots,w_5) = (1.0,\ 0.5,\ 1.5,\ 0.8,\ 0.3)$ are normalized weights empirically derived from a development set to reflect the practical importance of maximizing coverage and novelty while keeping scope narrowing and uncertainty in check. Each component is computed at the element level and then averaged: coverage rewards turning \textit{Disclosed} points into \textit{NotDisclosed}; scope penalises unnecessary narrowing; novelty counts only those changes the examiner rates as innovative; consistency multiplies legal readability with technical coherence; and uncertainty, derived from epistemic variance, discourages speculative edits.

To isolate genuine model doubt from data noise, total variance is decomposed into epistemic and aleatoric terms, $\sigma^{\text{total}}=\sigma^{\text{epi}}+\sigma^{\text{ale}}$ with $\sigma^{\text{ale}}=1-\text{confidence}$.  Only $\sigma^{\text{epi}}$ enters \(R_{\text{uncertainty}}\), so stochastic prior‑art phrasing does not unduly penalise promising revisions.

Finally, progressive widening controls the branching factor as the search deepens:
\begin{equation}
K(n) = \bigl\lceil \alpha N(n)^{\delta} \bigr\rceil,
\qquad (\alpha, \delta) = (2.0, 0.5)
\end{equation}
which encourages broad exploration early on and fine‑grained exploitation later without letting the tree explode.

\section{Experiments}
We conduct comprehensive experiments to evaluate the effectiveness, robustness, and interpretability of the proposed ToC framework in real-world patent claim editing scenarios. Our evaluation focuses on three core questions: 
\begin{itemize}
    \item Effectiveness: Does ToC generate higher-quality claims compared to strong LLM-based baselines, including zero-shot and few-shot prompting setups?
    \item Component Contribution: How do modules such as multi-agent collaboration, uncertainty gating, and progressive widening affect performance?
    \item Control and Stability: How sensitive is ToC to key hyperparameters like reward weights and uncertainty thresholds?
\end{itemize}

To answer these questions, we perform comparative evaluations, ablation studies, sensitivity analyses, and expert assessments. Results are measured across multiple objective and subjective dimensions, including novelty, scope preservation, legal readability, and human preference. The following subsections detail the experimental setup, dataset construction, evaluation metrics, and a performance analysis.

\begin{table*}[t]
\centering
\small
\begin{tabular}{llcccc}
\toprule
\textbf{Model} & \textbf{ToC} & JSON & PPL & ROUGE‑L & BLEU \\
\midrule
\multicolumn{6}{l}{\textbf{Closed‑Source MLLMs}}\\
OpenAI O1                 & \ding{51} & 0.994±0.003 & 9.10±1.10 & 0.602±0.050 & 0.540±0.051 \\
\hspace{2.7em}+ zero‑shot & \ding{55} & 0.992±0.004 & 9.42±1.15 & 0.567±0.050 & 0.497±0.049 \\
\hspace{2.7em}+ few‑shot  & \ding{55} & 0.993±0.004 & 9.25±1.14 & 0.587±0.049 & 0.518±0.050 \\[2pt]
GPT‑4o                    & \ding{51} & \textbf{0.996±0.001} & \textbf{8.72±1.00} & \textbf{0.624±0.049} & \textbf{0.554±0.052} \\
\hspace{2.7em}+ zero‑shot & \ding{55} & 0.994±0.003 & 8.93±1.05 & 0.590±0.046 & 0.522±0.050 \\
\hspace{2.7em}+ few‑shot  & \ding{55} & 0.995±0.003 & 8.85±1.03 & 0.610±0.048 & 0.537±0.051 \\[2pt]
Claude‑3.5 Sonnet         & \ding{51} & 0.995±0.003 & 8.98±1.08 & 0.611±0.050 & 0.530±0.051 \\
\hspace{2.7em}+ zero‑shot & \ding{55} & 0.993±0.004 & 9.30±1.14 & 0.582±0.049 & 0.501±0.050 \\
\hspace{2.7em}+ few‑shot  & \ding{55} & 0.994±0.002 & 9.18±1.10 & 0.595±0.049 & 0.513±0.050 \\[2pt]
\midrule
\multicolumn{6}{l}{\textbf{Open‑Source MLLMs}}\\
Qwen2.5‑VL‑32B            & \ding{51} & 0.992±0.004 & 9.80±1.20 & 0.582±0.050 & 0.510±0.051 \\
\hspace{2.7em}+ zero‑shot & \ding{55} & 0.990±0.005 & 10.08±1.25 & 0.551±0.050 & 0.478±0.050 \\
\hspace{2.7em}+ few‑shot  & \ding{55} & 0.991±0.005 & 9.96±1.23 & 0.567±0.050 & 0.490±0.050 \\[2pt]
Qwen2.5‑VL‑72B            & \ding{51} & 0.993±0.004 & 9.52±1.17 & 0.596±0.050 & 0.525±0.051 \\
\hspace{2.7em}+ zero‑shot & \ding{55} & 0.991±0.003 & 9.76±1.22 & 0.563±0.050 & 0.492±0.050 \\
\hspace{2.7em}+ few‑shot  & \ding{55} & 0.992±0.004 & 9.65±1.18 & 0.582±0.050 & 0.508±0.051 \\
\bottomrule
\end{tabular}
\caption{Auxiliary generation metrics for the same zero‑/few‑shot and ToC runs.}
\label{tab:llm_aux_metrics}
\end{table*}

\subsection{Experimental Setup}
All experiments are conducted on a high-performance computing cluster equipped with NVIDIA A100 GPUs (80GB). To ensure reproducibility, we fix all random seeds and repeat each experiment three times, reporting the mean and standard deviation. The system is implemented in Python, and all hyperparameters—including search depth, reward weights, and LLM decoding settings—are managed through a unified configuration interface.

We evaluate the ToC framework using a diverse set of multimodal LLMs as reasoning agents. For closed-source models, we include OpenAI O1 \cite{jaech2024openai}, GPT-4O \cite{hurst2024gpt}, and Claude 3.5 Sonnet \cite{claude2024sonnet}, accessed via their official APIs with deterministic decoding settings (temperature = 0.0). For open-source models, such as Qwen2.5-VL (32B/72B) \cite{wang2024qwen2}, all inference is performed locally using the HuggingFace Transformers library. Each model is deployed in both the examiner and editor roles within the ToC multi-agent loop, following consistent prompt structures and unified output schemas to ensure a fair and controlled comparison. All evaluations are conducted on standardized data splits with identical inputs and evaluation protocols across models.

\subsection{Dataset Preprocessing}
We construct our benchmark from the USPTO Office Actions Research Dataset \footnote{\url{https://data.uspto.gov/bulkdata/datasets/ptoffact?fileDataFromDate=2017-11-29&fileDataToDate=2017-11-29}}, focusing on wireless communications patents and 2016 office actions. The dataset contains 1,145 unique patents (106 allowed, 1,039 rejected), 28,261 claims, and 8,418 prior art references. 

The data preprocessing pipeline includes parsing structured patent and office action records, aligning examiner-cited prior art with rejected claims, mapping full claim and prior art texts, and decomposing claims into semantic sub-elements. Prior art documents are retrieved through automated web scraping from Google Patents using publication IDs extracted from office action records. Prior art documents are further processed to extract relevant evidence using embedding-based similarity filtering. Our dataset is multimodal in nature, incorporating both textual descriptions and supporting figures/diagrams from patent documents, enabling comprehensive analysis of technical content across different modalities. Each data instance is represented as a quadruple \( \langle c_i, y_i, e_i, r_i \rangle \), where \( c_i \) is a claim element, \( y_i \) is a binary disclosure label, \( e_i \) is the matched evidence (including both text and visual elements), and \( r_i \) is a justification. All annotations are reviewed by technical experts to ensure quality and consistency.

The dataset is imbalanced (9.3\% allowed, 90.7\% rejected), with allowed patents having more claims on average. 82.3\% of claims have prior art references. See Table~\ref{tab:dataset_detailed} for details.

\begin{table}[ht]
\centering
\small
\begin{tabular}{lcc}
\toprule
\textbf{Metric} & \textbf{Value} & \textbf{Percentage} \\
\midrule
Total Patents & 1,145 & 100\% \\
- Allowed Patents & 106 & 9.3\% \\
- Rejected Patents & 1,039 & 90.7\% \\
\midrule
Total Claims & 28,261 & 100\% \\
- Allowed Claims & 4,272 & 15.1\% \\
- Rejected Claims & 23,989 & 84.9\% \\
\midrule
Average Claims per Patent & 24.7 & - \\
\midrule
Patents with Prior Art & 574 & 50.1\% \\
Patents without Prior Art & 571 & 49.9\% \\
\midrule
Claims with Prior Art & 23,260 & 82.3\% \\
Claims without Prior Art & 5,001 & 17.7\% \\
\midrule
Prior Art References & 8,418 & - \\
Overlapping Applications & 27 & 2.4\% \\
\bottomrule
\end{tabular}
\caption{Detailed dataset description by category.}
\label{tab:dataset_detailed}
\end{table}

\subsection{Baselines and Evaluation Metrics}
To evaluate the effectiveness of the proposed ToC framework, we compare it against two baseline settings for each LLM: a zero-shot setting without any reasoning guidance, and a few-shot (2-shot) setting with exemplars and Chain-of-Thought reasoning prompts. All setups adopt identical input-output formats, prompts, and search budgets to ensure fair comparison across models and configurations.

We evaluate every model on a 500‑sample hold‑out set of granted and rejected claims.
Core metrics follow our reward terms: coverage F1 and \( \Delta R_\mathrm{coverage} \) for disclosure avoidance, scope similarity/loss for scope retention, novelty, consistency, and an uncertainty score. To probe general robustness we additionally report JSON‑parsing completeness, chain entropy, image–text consistency, legal readability, token‑level perplexity (PPL), as well as ROUGE‑L, BLEU, and expert preference.

\subsection{Experimental Results}
\subsubsection{Performance Comparison.}
Table \ref{tab:llm_primary_metrics} and table \ref{tab:llm_aux_metrics} show that integrating ToC markedly boosts every model, with the closed‑source GPT‑4o leading overall. It posts the best coverage (0.582), novelty (0.732), consistency (0.956), and lowest perplexity (8.72), giving it the highest aggregate score (0.701). Among open‑source options, Qwen2.5‑VL‑72B ranks first, outperforming its 32B variant on all reward‑aligned metrics while remaining within 5\% of the GPT‑4o benchmark. The strong gains of both Qwen models over their non‑ToC baselines confirm that the framework transfers reliably across parameter scales.

\begin{figure*}[t]
  \centering
  \begin{subfigure}[t]{0.25\textwidth}
    \centering
    \includegraphics[width=\linewidth]{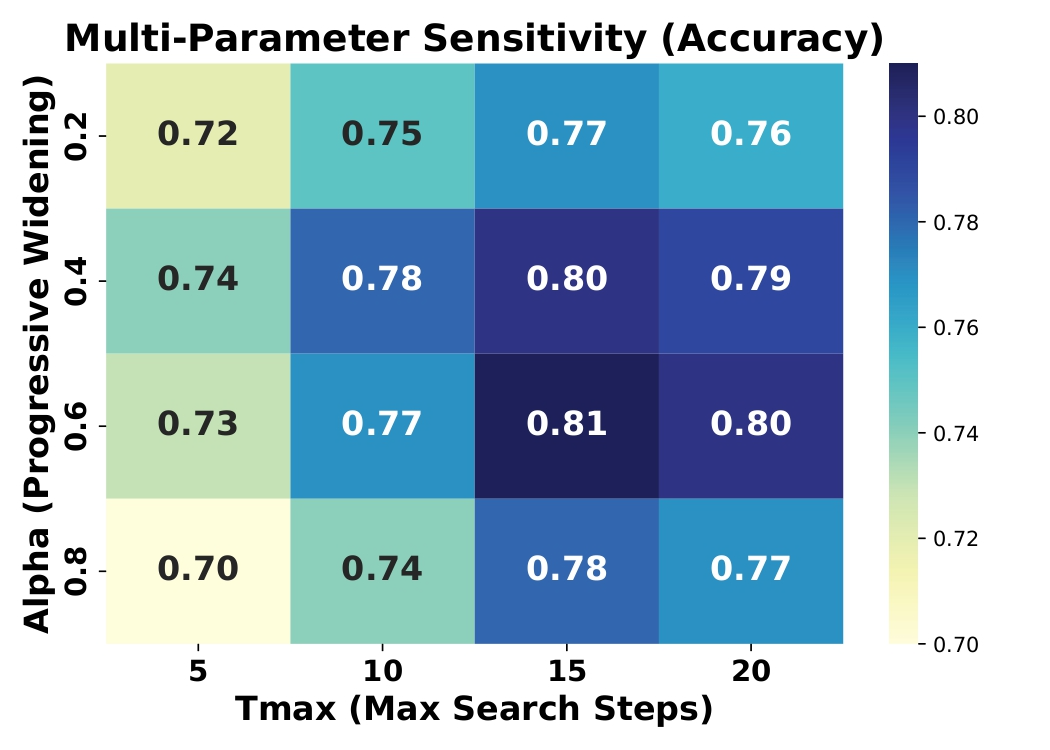}
    \caption{Sensitivity heat‑map: accuracy vs.\ uncertainty threshold and MCTS depth.}
    \label{fig:sensitivity}
  \end{subfigure}
  \hfill
  \begin{subfigure}[t]{0.25\textwidth}
    \centering
    \includegraphics[width=\linewidth]{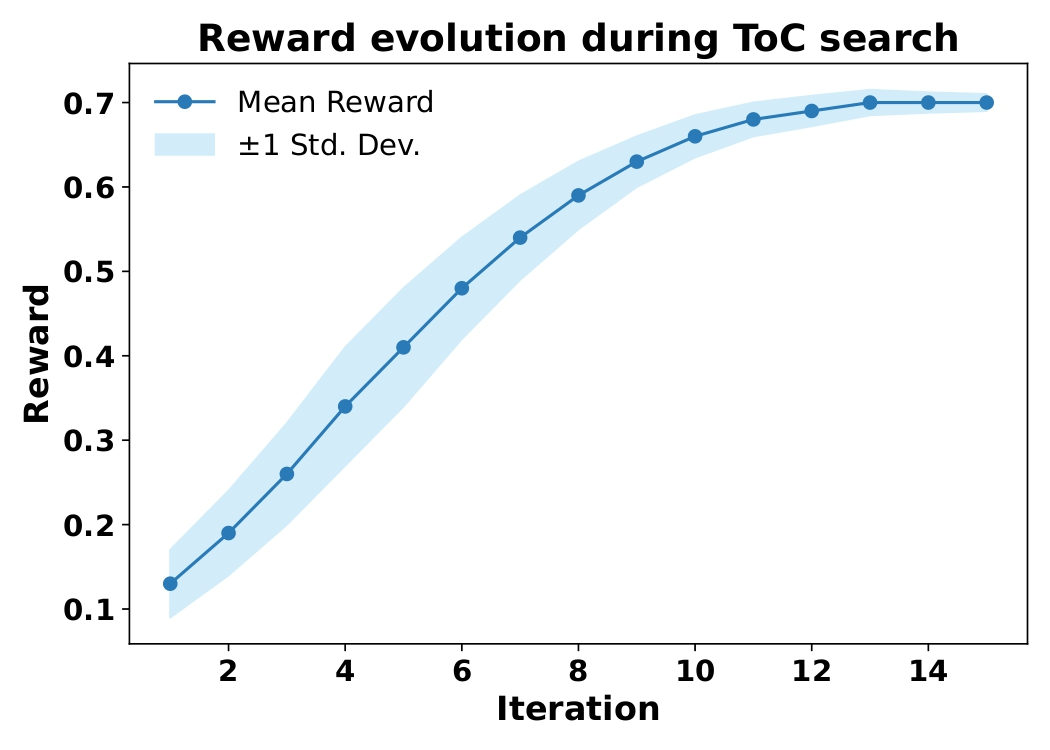}
    \caption{Reward evolution during the search.}
    \label{fig:reward_curve}
  \end{subfigure}
  \hfill
  \begin{subfigure}[t]{0.47\textwidth}
    \centering
    \includegraphics[width=\linewidth]{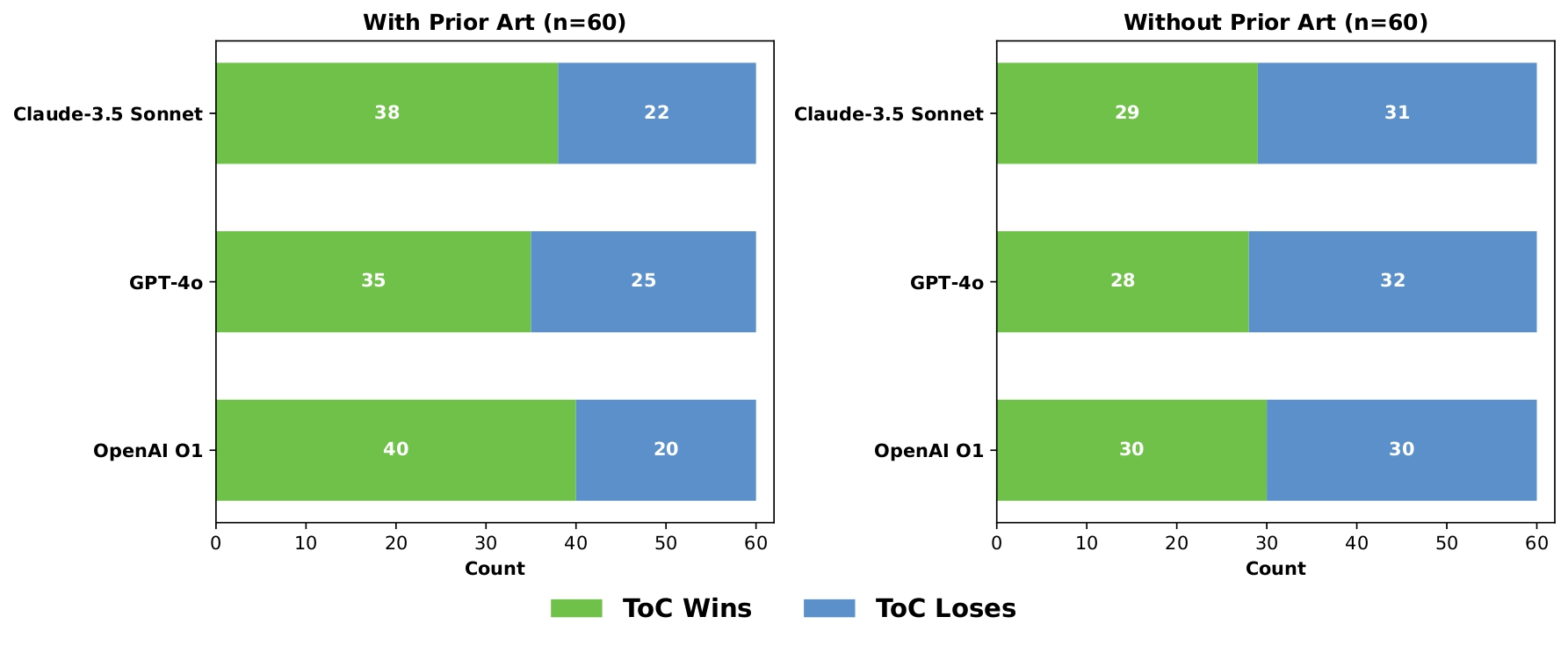}
    \caption{Expert preference scores for final claims.}   
    \label{fig:expert}
  \end{subfigure}
  \caption{Visual analysis of ToC optimisation.  
  (a) Sensitivity to hyper‑parameters.  
  (b) Reward dynamics across search steps.  
  (c) Human expert evaluation of generated claims.}       
  \label{fig:combined_figures}
\end{figure*}

\subsubsection{Error Analysis.}
Expert evaluations involved five patent specialists with at least five years of industry experience, evaluating revisions against a detailed rubric focusing on novelty, legal robustness, and technical coherence. Additionally, these experts conducted comparative evaluations between ToC-revised claims and original drafts, revealing a clear preference for ToC revisions in contexts where prior art was available. Specifically, ToC was preferred in approximately two-thirds of evaluations (see Figure~\ref{fig:expert}). The detailed error analysis (Table~\ref{tab:error_analysis}) identified prevalent \textit{System Control Failures}, notably including excessive branching due to uncertainty, zero-confidence decisions, missed pruning opportunities, and invalid modifications leading to semantic or legal inconsistencies. Further frequently encountered errors involved unsupported claims of novelty, highlighting areas for future improvement in uncertainty calibration, validation mechanisms, and alignment with prior-art evidence.

\begin{table}[t!]
\centering
\small
\begin{tabular}{lcc}
\toprule
\textbf{Error Category} & \textbf{Count} & \textbf{Proportion (\%)} \\
\midrule
Input Misalignment & 78 & 6.8 \\
Invalid Modifications & 120 & 10.5 \\
Unsupported Novelty & 113 & 9.9 \\
Legal or Style Issues & 31 & 2.7 \\
System Control Failures & 139 & 12.2 \\
\bottomrule
\end{tabular}
\caption{Error analysis: categorized failure types in ToC-generated revisions.}
\label{tab:error_analysis}
\end{table}

\subsubsection{Ablation Study.}
\begin{figure}[ht]
\centering
\includegraphics[width=0.65\linewidth]{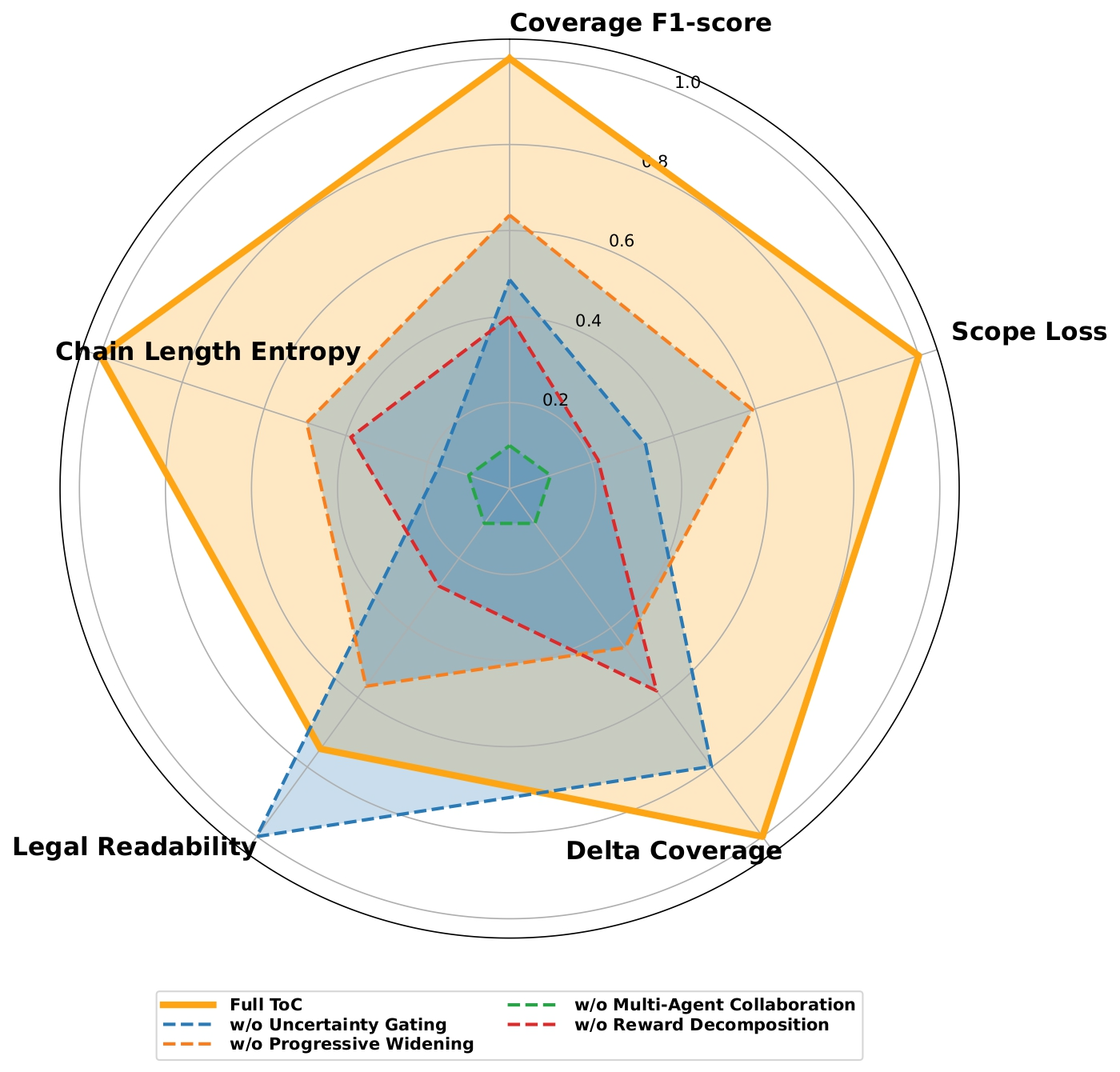}
\caption{Component-level performance under module ablations. Full ToC includes all modules; others have one component removed.}
\label{fig:ablation_study}
\end{figure}

To understand the contribution of each module within ToC, we conduct a comprehensive ablation analysis. As shown in Figure~\ref{fig:ablation_study}, disabling components such as uncertainty gating, progressive widening, or multi-agent collaboration results in noticeable performance degradation. In particular, the absence of uncertainty control and agent interaction significantly impacts novelty, coverage, and overall legal quality.

\subsubsection{Sensitivity Analysis.}
Figure \ref{fig:sensitivity} sweeps two key hyper‑parameters: the progressive‑widening coefficient $\alpha$ (rows) and the maximum search depth $T_{\max}$ (columns). Accuracy is remarkably stable (0.72–0.81) across the grid, peaking at $\alpha = 0.6$ and $T_{\max} = 15$, a setting that balances the breadth of candidate edits with sufficient roll‑out depth. Very small trees ($T_{\max} \le 5$) or overly aggressive widening ($\alpha = 0.8$) shave off 3–5 points, confirming that extreme exploration or shallow search can dilute reward optimisation. Figure \ref{fig:reward_curve} complements the heat‑map by tracking the mean reward over search iterations. Roughly 70\% of the eventual gain is achieved within the first six iterations; the curve then plateaus after iteration 10, indicating diminishing returns and validating our depth budget.

\section{Conclusion}
We introduced ToC (Tree-of-Claims), a novel framework combining structured search and multi-agent collaboration for patent claim optimization. By modeling editing as a sequential decision-making task via Monte Carlo Tree Search (MCTS), ToC provides a transparent and controlled approach that significantly improves novelty, legal quality, and expert preference compared to existing methods.

ToC represents a paradigm shift from passive generation to strategic collaboration, enhancing correctness, transparency, and interpretability in high-stakes legal writing tasks and facilitating effective human-AI interaction.

Future work will focus on incorporating multimodal reasoning for figure-grounded edits, improving search efficiency through distributed computing, and generalizing the framework to other structured editing domains such as legal contracts, medical protocols, and scientific methods. Compared to traditional pipelines like structured prompt chaining or memory-based editing, ToC offers stronger control, traceability, and multi-turn reasoning, making it particularly suited for domains requiring high reliability and domain-specific consistency.

\bibliography{aaai2026}

\clearpage

\setcounter{section}{0}
\renewcommand\thesection{\Alph{section}} 

\section{Case Study}
Table~\ref{tab:case_study} illustrates a real-world example of ToC editing a claim step-by-step. Each operation is guided by the ExaminerAgent's feedback and strategically chosen by the EditorAgent. The final claim achieves improved novelty and passes the disclosure check, demonstrating the framework's practical utility.

\begin{table}[ht!]
\centering
\small
\caption{Case study: Step-wise modification of a claim by the ToC system.}
\label{tab:case_study}
\renewcommand{\arraystretch}{1.2}
\begin{tabularx}{\columnwidth}{@{}p{2.2cm}X@{}}
\toprule
\textbf{Step - Operation} & \textbf{Claim Content and Reasoning} \\
\midrule
Original Claim &
\textbf{Claim:} A system for image processing, comprising: a processor configured to receive an input image and apply a filter to the image. \\
\midrule
1 - AddNovelFeature &
\textbf{Modified:} A system for image processing, comprising: a processor configured to receive an input image, apply a filter to the image, \textcolor{blue}{\textbf{and perform adaptive contrast enhancement}}. \newline
\textbf{Examiner:} The added enhancement feature is not disclosed in prior art. \newline
\textbf{Editor:} Introduce novel technical content to improve distinctiveness. \\
\midrule
2 - ReplaceSynonym &
\textbf{Modified:} A system for image processing, comprising: a processor configured to receive an input image, \textcolor{blue}{\textbf{process the image}}, and perform adaptive contrast enhancement. \newline
\textbf{Examiner:} Broader phrasing still supports novelty with enhancement. \newline
\textbf{Editor:} Improve clarity and broaden scope using alternate terminology. \\
\midrule
3 - AddLimitation &
\textbf{Modified:} A system for image processing, comprising: a processor configured to receive an input image, process the image, and perform adaptive contrast enhancement, \textcolor{blue}{\textbf{wherein the enhancement is performed only on grayscale images}}. \newline
\textbf{Examiner:} Specific constraint not disclosed and enhances novelty. \newline
\textbf{Editor:} Restrict technical scope to reinforce inventive distinction. \\
\midrule
Final Claim &
\textbf{Claim:} A system for image processing, comprising: a processor configured to receive an input image and perform adaptive contrast enhancement, wherein the enhancement is performed only on grayscale images. \newline
\textbf{Outcome:} The final version passes disclosure check with improved novelty and clarity. \\
\bottomrule
\end{tabularx}
\end{table}

\section{Prompt Templates} 

\subsection{Global Configuration \& Schema.}

\paragraph{Global Parser Contract.}
\begin{quote}
    \small \ttfamily 
    \begin{itemize}
        \setlength{\itemsep}{0pt}
        \setlength{\parskip}{0pt}
        \setlength{\leftmargin}{1em}
        \item Simulation policy: "Hybrid" (entropy/confidence = 0.6/0.4).
        \item $\sigma$-epi gating threshold: 0.20; human\_review := (uncertainty $>$ 0.20).
        \item Progressive widening: $K(n) = \lceil \alpha \cdot N(n)^\delta \rceil$ (fixed $\alpha, \delta$).
        \item All prompts must output STRICT JSON; no text outside JSON.
    \end{itemize}
\end{quote}

\paragraph{Micro-Validation Schema.}
\begin{quote}
    \small \ttfamily 
    \vspace{0.5em}
    \begin{itemize}
        \setlength{\itemsep}{0pt}
        \setlength{\parskip}{0pt}
        \setlength{\leftmargin}{1em}
        \item \textbf{Examiner JSON Keys}: ["status", "evidence\_text", "reasoning", "confidence", "uncertainty", "human\_review"]
        \item status $\in$ \{"Disclosed", "PartiallyDisclosed", "NotDisclosed"\}
        \item confidence, uncertainty $\in$ [0.00, 1.00] (2 decimals)
        \item \textbf{Editor JSON Keys}: ["operations"]; non-empty list of objects:
        \begin{itemize}
            \setlength{\itemsep}{0pt} \setlength{\leftmargin}{1em}
            \item operation\_type $\in$ \{AddNovelFeature, \dots, AddDependency\}
            \item target\_element\_id, modified\_text, rationale: string
            \item confidence: float (2 decimals)
        \end{itemize}
    \end{itemize}
\end{quote}

\subsection{ExaminerAgent Prompts \& Schema} 

\paragraph{System Message.}

\begin{quote}
    \small\ttfamily
    You are a professional patent examiner with extensive experience in patent examination. Your tasks are:
    \begin{enumerate}
        \item Carefully analyze the technical features of the claim
        \item Search for corresponding technical content in the prior art
        \item Determine whether the claim is disclosed by the prior art
        \item Provide a detailed reasoning process and evidence
        \item Assess the confidence of the examination result
    \end{enumerate}
    Please strictly provide your response in the required JSON format.
\end{quote}

\paragraph{Main Prompt (per element).}

\begin{quote}
    \small\ttfamily
    As a professional patent examiner, please determine whether the following claim element is disclosed by the given prior art document. Based on the content of the prior art and your professional judgment, output a structured conclusion.
    
    \vspace{0.5em}
    {[Claim Element]} \\
    ID: \{claim\_element.element\_id\} \\
    Type: \{claim\_element.element\_type\} \\
    Text: \{claim\_element.text\}
    
    \vspace{0.5em}
    {[Prior Art Content]} \\
    \{prior\_art.description\}
    
\vspace{0.5em}
STRICTLY output the following JSON (no markdown, no backticks): \\
\{ \\
\hspace*{1em} "status": "Disclosed|PartiallyDisclosed| \\
\hspace*{5em} NotDisclosed", \\
\hspace*{1em} "evidence\_text": "verbatim quote(s) from prior \\
\hspace*{5em} art or 'None'", \\
\hspace*{1em} "reasoning": "concise reasoning ($<$120 words) \\
\hspace*{5em} with equivalence mapping", \\
\hspace*{1em} "confidence": 0.00, \\
\hspace*{1em} "uncertainty": 0.00, \\
\hspace*{1em} "human\_review": false \\
\}

\vspace{0.5em}
Notes: \\
- "status" is a single label (case-sensitive). \\
- "confidence" $\in$ [0.00, 1.00] . \\
- "uncertainty" $\in$ [0.00, 1.00] and denotes epistemic variance (K stochastic runs). \\
- Set "human\_review": true iff uncertainty $> 0.20$ ($\sigma$-epi gating threshold). \\
- Use exact quotes for "evidence\_text" when available; otherwise "None". \\
- Output only valid JSON; double quotes; floats with 2 decimals.
    
    Ensure the output is complete, accurate, and professional.
\end{quote}

\subsection{EditorAgent Prompts \& Schema} \paragraph{System Message.}

\begin{quote}
    \small\ttfamily
    You are a professional patent attorney with extensive experience in patent drafting and amendment. Your tasks are:
    \begin{enumerate}
        \setlength{\itemsep}{0pt}
        \setlength{\parskip}{0pt}
        \item Analyze the disclosure status of claim elements
        \item Select appropriate edit operations
        \item Modify the claim to avoid being disclosed by the prior art
        \item Maintain the integrity and feasibility of the technical solution
        \item Preserve an appropriate scope of protection
    \end{enumerate}
    Please strictly provide your response in the required JSON format.
\end{quote}

\paragraph{Edit Planning Prompt (per element).}

\begin{quote}
    \small\ttfamily
    Please modify the following disclosed claim element to avoid being disclosed by the prior art:
    
    \vspace{0.5em}
    Original Claim Element: \\
    ID: \{element.element\_id\} \\
    Type: \{element.element\_type\} \\
    Text: \{element.text\}
    
    \vspace{0.5em}
    Disclosure Information: \\
    Status: \{reasoning\_chain.status\} \\
    Evidence: \{reasoning\_chain.evidence\_text\} \\
    Reasoning: \{reasoning\_chain.reasoning\}
    
    \vspace{0.5em}
    Available Edit Operations:
    \begin{itemize}
        \setlength{\itemsep}{0pt}
        \setlength{\parskip}{0pt}
        \setlength{\leftmargin}{1em}
        \item \footnotesize AddNovelFeature
        \item \footnotesize ReplaceSynonym
        \item \footnotesize ReframeViaFigure
        \item \footnotesize DropElement
        \item \footnotesize MergeElements
        \item \footnotesize SplitElement
        \item \footnotesize AddLimitation
        \item \footnotesize ModifyRelationship
        \item \footnotesize ChangeOrder
        \item \footnotesize AddDependency
    \end{itemize}

    \vspace{0.5em}
    STRICTLY output the following JSON: \\
    \{ \\
    \hspace*{1em} "operations": [ \\
    \hspace*{2em} \{ \\
    \hspace*{3em} "operation\_type": "$<$one of the Allowed \\
    \hspace*{5em} Atomic Operations$>$", \\
    \hspace*{3em} "target\_element\_id": "\{element.element\_id\}", \\
    \hspace*{3em} "modified\_text": "revised text for this \\
    \hspace*{5em} element only", \\
    \hspace*{3em} "rationale": "how it breaks mapped evidence \\
    \hspace*{5em} while preserving scope", \\
    \hspace*{3em} "confidence": 0.00 \\
    \hspace*{2em} \} \\
    \hspace*{1em} ] \\
    \}
    
    \vspace{0.5em}
    Rules: \\
    - Use only the enumerated operation\_type values (exact spellings). \\
    - "modified\_text" must be legally styled, non-trivial, and feasible. \\
    - Prefer minimal change that defeats the cited evidence. \\
    - confidence $\in$ [0.00, 1.00], 2 decimals. No extra keys.
\end{quote}

\paragraph{Apply-Specific-Operation Prompt (single operation).}

\begin{quote}
    \small\ttfamily
    Please apply the specified edit operation to modify the claim element:
    
    \vspace{0.5em}
    Original Element: \\
    ID: \{element.element\_id\} \\
    Text: \{element.text\}
    
    \vspace{0.5em}
    Disclosure Information: \\
    \{reasoning\_chain.reasoning\}
    
    \vspace{0.5em}
    Edit Operation Type: \{operation\_type\}
    
    \vspace{0.5em}
    Please provide the modified text and reasoning in the following format: \\
    \{ \\
    \hspace*{1em} "modified\_text": "Modified text", \\
    \hspace*{1em} "reasoning": "Reason for modification", \\
    \hspace*{1em} "confidence": 0.85 \\
    \}
\end{quote}

\end{document}